\documentclass[runningheads]{llncs}

\usepackage{graphicx}
\usepackage{algorithm}
\usepackage{enumitem}
\PassOptionsToPackage{hyphens}{url}\usepackage{hyperref}

\usepackage[noend]{algpseudocode}
\usepackage{type1cm}      
\usepackage{graphicx}        
\usepackage[bottom]{footmisc}
\usepackage{newtxmath}      
\usepackage{color}
                
\usepackage[numbers,sectionbib]{natbib}   
\usepackage{enumitem}
\usepackage[T1]{fontenc}
\usepackage{xspace}

\usepackage{caption}
\usepackage{subcaption}

\usepackage[framemethod=TikZ]{mdframed}
\usepackage{wrapfig}

\newcommand{\vadalog}{\textsc{Vadalog}\xspace}
\newcommand{\ebaldone}{$\epsilon$-Baldone\xspace}
\newcommand{\econglomerate}{$\epsilon$-conglomerate\xspace}

\begin{document}
\title{COVID-19 and Company Knowledge Graphs: \\Assessing Golden Powers and Economic Impact \\ of Selective Lockdown via AI Reasoning\thanks{The views and opinions expressed in this paper are those of the authors and do not necessarily reflect the official policy or position of Banca d'Italia. This version is dated 2020-04-21.}}
\titlerunning{COVID-19 and Company Knowledge Graphs}
\author{Luigi Bellomarini\inst{1}
 \and
Marco Benedetti\inst{1} \and
Andrea Gentili\inst{1} \and
Rosario Laurendi\inst{1} \and
Davide Magnanimi\inst{1}\and 
Antonio Muci\inst{1} \and
Emanuel Sallinger\inst{2,3}
}
\authorrunning{L. Bellomarini et al.}
\institute{Banca d'Italia \and
TU Wien \and
University of Oxford
}
\maketitle

\begin{abstract}
In the COVID-19 outbreak, governments have applied progressive restrictions to production activities, permitting only those that are considered strategic or that provide essential services. This is particularly apparent in countries that have been stricken hard by the virus, with Italy being a major example. Yet we know that companies are not just isolated entities: They organize themselves into intricate shareholding structures --- forming company networks --- distributing decision power and dividends in sophisticated schemes for various purposes.

One tool from the Artificial Intelligence (AI) toolbox that is particularly effective to perform reasoning tasks on domains characterized by many entities highly interconnected with one another is Knowledge Graphs (KG). In this work, we present a visionary opinion and report on ongoing work about the application of Automated Reasoning and Knowledge Graph technology to address the impact of the COVID-19 outbreak on the network of Italian companies and support the application of legal instruments for the protection of strategic companies from takeovers.
\end{abstract}

\section*{Summary}

In the COVID-19 outbreak, governments have applied progressive restrictions to production activities,
permitting only those that are considered strategic or that provide essential services.
This is particularly apparent in countries that have been stricken hard by the virus, with
Italy being a major example. In the case of Italy, the Government has been issuing multiple
decrees (three as of this writing), often amended or integrated with local provisions, specifying the
allowed economic sectors for each emergency period. Most other countries are releasing similar regulations, with different frequencies and time spans.

\medskip
\noindent
\textbf{Challenges}.
In the global economic system, companies are structured in complex shareholding structures, 
decoupling the so-called \emph{integrated ownership}, i.e., the actual rights on the dividends of a company, 
from \emph{company control}, which corresponds to the possibility of the shareholders to 
take decisions by controlling the vote majority. The combination of all the shareholding structures of
the companies in one country constitutes a company graph or company network, a topology
where shareholders (companies and people) can be seen as nodes and ownerships as links,
connecting those nodes. In such structures, understanding the real influence of a company over another in terms of
ownership and control is not trivial, and requires sophisticated
analysis on the overall network that cannot be accomplished with ordinary data management technology.

\medskip
Conglomerates are groups of companies that are in some sense close in the company graph.
A conglomerate or a company group is a cluster containing companies
with shared financial interests or representing the same center of interest or power.
Conglomerates can be extremely small, counting from two up to a few tens of companies, 
or very large, with hundreds or thousands of entities. When crisis strikes, 
the presence of groups is a further chance for companies to survive, thanks to the resilience deriving from 
geographical differentiation, diversification of economic activities, improved risk sharing, and so on.
Yet, lockdown alters mutual influence between companies and undermines 
the structure of conglomerates; sometimes companies are 
even legally impeded from paying their dividends. Similarly, control relationships
and actual decision centers become hard to individuate.
Not to mention that the topology of the network can change
relatively quickly over time, with small mutations possibly yielding large cascade effects, on which a swift and total re-assessment is to be made.

\medskip
A crisis induces even more dynamics on the company network, if we consider strategic enterprises.
They are specific entities (e.g., in the energy, military, transport fields, etc.) that are considered of national
relevance and must therefore be carefully protected in terms of shareholding structure, so they remain
publicly held, i.e., with the State controlling the majority of the shares. 
It turns out that in crises, taking advantage of market turbulence, specific players are inclined to pursuing
takeovers and affect the public control over such companies. This theme has received 
prominent attention with the COVID-19 outbreak and is especially relevant in the Italian case.
Many countries have developed legal frameworks to protect strategic companies by vetoing 
specific share acquisition operations, which are likely to hide a takeover attempt. Yet, the problem of detecting when
a specific transaction underlies a takeover is by no means trivial and
involves a global approach which considers at least the entire national company graph.

\medskip
\noindent
\textbf{General Aim of the Work}.
We present a visionary opinion and report on ongoing work about the application of Automated Reasoning and
Knowledge Graph technology to address the impact of the COVID-19 outbreak on the network of Italian companies and support the
application of legal instruments for
the protection of strategic companies from takeovers.

\begin{itemize}[leftmargin=2mm,noitemsep]
\item Our mid-term goal is contributing summary
information as well as data (in the form of Knowledge Graphs) to aid specialists, analysts and decision makers
to design the best solutions for supporting businesses and policy-making.
\item  We aim at providing
methodologies of general validity, relevant in any economic turbulence. We start and
operate with special reference to the Italian COVID-19 outbreak. 
\end{itemize}

\medskip\noindent\textbf{Automated Reasoning and Knowledge Graphs}.
 Knowledge Graphs are a state-of-the-art tool from the Artificial Intelligence toolbox,
particularly effective to model and perform automated reasoning tasks on domains characterized
by the presence of a big volume of highly interconnected entities. Indeed, reasoning is the process of
deducing new information for the available networks given a formal representation
of the domain. Logic-based languages for Knowledge Representation and
Reasoning are a largely adopted tool, enjoying recently increased popularity thanks to the rise of
mature formalisms. They strike a good balance between expressive power and computational complexity,
in the sense that they are sufficiently articulated to be able to represent real-world problems
and simple enough to be efficiently processed by automated reasoning systems for Knowledge Graphs
(Knowledge Graph Management Systems --- KGMS).
We have specific experience with Vadalog, a state-of-the-art KGMS we have been developing with the
University of Oxford. We have already put it to good use while reasoning on large economic networks of interest to the Bank of Italy.

\medskip\noindent\textbf{Contribution}.
In this work, we focus on methodological and technical Artificial Intelligence  tools to evaluate the resilience of the Italian company network,
with special attention to conglomerates.  We use Vadalog and Knowledge Graphs in general, to provide a reference
modeling framework for the problems at hand.

\begin{itemize}[leftmargin=2mm,noitemsep]
\item We propose a novel and general formalization of the notion of integrated ownership, which we
then use to introduce our concept of conglomerate, a set of companies that are ``close'' w.r.t. integrated ownership.
\item We present and discuss the different Knowledge Graphs we built, referring to the Italian company 
network before the COVID-19 outbreak and after the release of each Italian shutdown decree. 
\item For these Knowledge Graphs, we provide standard network analytics as well as
complex ones, based on a principled combination of conglomerates and integrated ownership.
\item We study the problem of protecting strategic companies
from takeovers and propose a formalization of the most relevant scenarios, opening to
deeper research and the development of dedicated services to support decision makers in the
evaluation of transactions involving company shares.
\end{itemize}

\medskip
\noindent
This work is indeed a working paper, collecting and presenting increasingly comprehensive analyses and results,
produced concurrently with the evolution of the outbreak.
All the produced Knowledge Graphs and indicators could be made available upon request and specific agreement.

\setcitestyle{square}

\section{Introduction}

With the COVID-19 outbreak, governments are applying progressive restrictions (\emph{shut down}) on the allowed economic activities, permitting only those that are considered strategic, while others are unavoidably screeching to a halt, possibly dragging countries into recession.

\medskip
Yet we know that companies are not just isolated entities: They organize themselves into intricate shareholding structures
---~forming \emph{company networks}~--- distributing decision power and dividends in sophisticated schemes for various purposes, not the least of which is, quite pertinently, resilience. In this respect, recent literature~\cite{Barca01,Franks95,Porta99} shows that \emph{concentrated corporate ownership} is
very common in industrialised countries, where a few subjects (or groups thereof) exert control over
a range of firms and favour the formation of conglomerates, i.e., company groups revolving around
a single center of interest or decision. In this context, convoluted and sometimes impenetrable ownership structures are being more and more used
as a tool to separate \emph{corporate ownership}, which pertains to share possession and hence to dividend flows,
from \emph{control}, which is the ability to push decisions through in the company and is related to voting power~\cite{ABIS20,Glat10}.

\medskip\noindent\textbf{Conglomerates}.
Chances for company conglomerates to overcome crises grow if they managed (long before it became necessary) to differentiate by specialization, that is, supply chain segment, 
economic activity, geography and, at the same time, if they got adaptive and thus ready to establish new connections and prune off dead branches, forming new more resilient groups or consolidating existing ones. 
They must distribute and mitigate default, credit, and operational risks, by means of a proper balancing of their shareholding structure. 
So, we ask: What are the dynamics induced by emergency restrictions on networks of companies? How are these networks going to adapt as (and if) conditions get stricter or more lenient over time?
And finally: Who benefits from all of this? It is clear that with the emergencial reorganizations in place, voting power and in general the ability to push decisions through in companies shift around: But in whose hands? 

\medskip\noindent\textbf{Strategic Companies}. Companies are key to the resilience of a state as a whole and it is essential that the ownership of --- or, more importantly, the decision power 
upon --- companies deemed of strategic relevance (e.g., in the energy, military, transport, telecommunications sectors) remains inside the national borders. 
Conversely, it is often the case that companies stretched by massive shut downs like those of COVID times 
(and production plunge in general) are subject to attempts of foreign takeovers, with buyers trying to take advantage of lowered share prices due to market uncertainty.
Different countries have classically resorted to legal frameworks 
to protect strategic companies against foreign takeovers~\cite{KiSh12}. Italy is a relevant example:
Being one of the most stricken countries, in the COVID emergency, it carried out a careful application of the so-called
\emph{Golden Powers}~\cite{reuters1}, that is the possibility of the central Government to veto
specific acquisition transactions (e.g., in terms of shares of stocks), that would cause strategic firms to be subject to
foreign takeover. Likewise, the Government can intervene to secure companies by acquiring or
increasing its participation (technically, \emph{investment beef-up}) via publicly controlled intermediaries
on the strategic firms.
Unfortunately, an effective application of the mentioned legal frameworks and Golden Powers in particular is by no means trivial.
How can we tell whether a transaction is a takeover attempt? Will a transaction actually allow a takeover?
What is the minimum amount of share that must shift to public control in order to protect a company?
And, how to protect against coordinated, \emph{collusive}, transactions aiming at a takeover?
Also, the transaction alone is not enough. In this game the corporate structure is essential: 
It can be either intrinsically prone to takeovers, for example
if there are firms acting as single points of attack so that gaining control over them
immediately allows for many takeovers, or resilient, for example if the control is reasonably shared
amongst multiple actors.

\medskip\noindent\textbf{Knowledge Graphs}. One tool from the Artificial Intelligence (AI)
toolbox that is particularly effective to perform reasoning tasks on domains
characterized by many entities highly interconnected with one another is
Knowledge Graphs (KG). Although there is no agreement on a single definition~\cite{EhrlingerW16},
KGs can be seen as semi-structured data models, able to represent knowledge about domains of interest
in the form of \emph{extensional facts} --- nodes and edges along with their properties --- and
\emph{intensional knowledge}, expressed in some formalism, for example logical rules.
Intensional knowledge is then activated on the extensional facts in the so-called \emph{reasoning process} 
so as to generate new knowledge in the form of new nodes and new edges~\cite{BFGS19}.\footnote{New nodes and
edges in this setting are not, of course, new companies
and links, but other objects/relationships such as
company groups, control connections, etc.} 
In simpler terms, not only do KGs offer an effective representation of highly interconnected data,
but these representations are also semantic, meaning that they allow algorithms to simulate a form of human cognition on top of the modeled objects, i.e., 
to perform \emph{Automated Reasoning} (AR), typically seen as the ideal complement
to machine learning within the vast field of AI~\cite{BGPS17,chowdury}.

\medskip
We have been successfully developing state-of-the-art AR tools on KGs, namely, 
\emph{Knowledge Graph Management Systems} (KGMS)~\cite{BeSG18}
and applying them to large networks of economic entities in the Bank of Italy~\cite{ABIS20}  with the goal of enhancing
the cognitive abilities of human analysts.
Although some recent initiatives~\cite{ATLAS} already attempt to track, analyze, and visualize market dynamics over time with a network-based approach, they specifically concentrate on trade flows across markets, and do not consider either company conglomerates or the protection of strategic companies. 
We rather believe KGs and AR can be effectively put to the disposal of the contingency, 
and contribute to a thorough inspection of the ongoing processes in company networks.

\medskip\noindent\textbf{Contribution}. This paper presents a visionary opinion about the application of AR and KG technologies 
to impact analysis of the COVID-19 outbreak on the network of Italian companies. In particular,
we concentrate on methodological and technical tools to evaluate
conglomerate resilience as well as protection measures for the strategic companies.
Our mid-term goal is contributing summary information to aid specialists (economists, analysts, decision makers)
to design the best solutions to support businesses and for policymaking. While we explicitly
refer to the current emergency with special respect to the Italian case and will refer to data
regarding such situation, we aim at providing methodologies that are valid in general,
independently of the country and if the specific crisis situation.
Indeed, assessing conglomerate resilience and protecting national strategic companies
is relevant in any situation of market turbulence, where production slow-down and foreign takeover attempts are typically involved.

\medskip\noindent
In particular, this paper contributes the following:

\begin{itemize}[leftmargin=2mm]

\item A formal \emph{definition of company conglomerate} suitable for
reasoning on their resilience with KGs.

\item The \emph{reference to a set of KGs} modeling the network of Italian companies in various configurations:
in the standard scenario before the COVID-19 outbreak and after
the issuance of several Government decrees, each introducing different limitations to production.
The full datasets of the KGs discussed in the paper is made available upon specific request and agreements.
\item A description of the characteristics of the Italian network of companies
in terms of \emph{standard network analytics} in the above configurations.
\item A description of the characteristics of the Italian network of companies in terms of
\emph{analytics specifically developed to estimate the resilience of conglomerates} in the above configurations.
\item A formal characterization of a set of \emph{reasoning tasks about Golden Powers}, 
modeling the possibility of different Governments to intervene on acquisition transactions that may
hide takeover attempts. In particular, we provide declarative specifications that allow for the development
of dedicated services aiming at deciding on vetoing transactions that may lead to takeovers or
giving insights about public investments that need to be increased in order to preserve specific companies.
Our technique considers non-local effects of transactions and handles collusive takeovers.

\end{itemize}

\medskip\noindent\textbf{Overview}. The remainder of the paper is organized as follows.
Section~\ref{sec:kgs} provides an introduction to the Knowledge Graph technology.
Section~\ref{sec:italian_db} presents the standard analytics of the Italian network,
before the COVID-19 outbreak. Section~\ref{sec:models} introduces our models
for company ownership and control. Then, in Section~\ref{sec:ateco} we present
the standard analytics for the shutdown configurations of the company network.
We present our model for conglomerates in Section~\ref{sec:conglomerates} together
with the respective analyses on the configurations at hand. Section~\ref{sec:golden_power}
shows our model for Golden Power and the construction of related added value services.
Section~\ref{sec:relwork} discusses the related work, and Section~\ref{sec:conclusions}
concludes the paper.

\section{Knowledge Graphs}
\label{sec:kgs}

\begin{wrapfigure}{R}{0.5\textwidth}
\centering
    \includegraphics[scale=0.3]{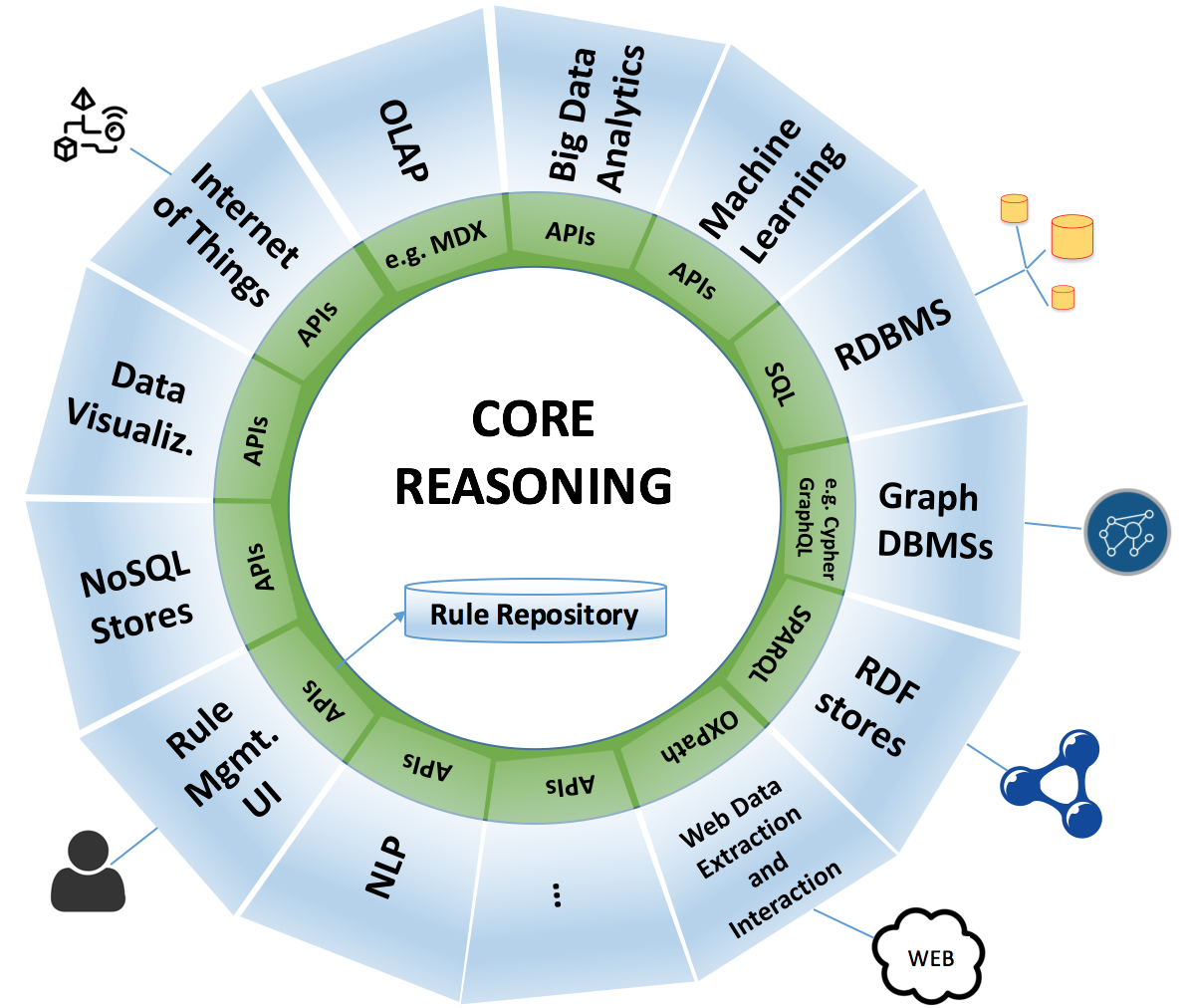} 
\vspace{-3mm}
\caption{\small{KGMS Reference Architecture.}}
\vspace{-7mm}
\label{fig:architecture}
\end{wrapfigure}

More and more companies wish to maintain knowledge in the form of an enterprise knowledge graph~\cite{wiki:kg} and to use and manage this knowledge via 
\emph{Knowledge Graph Management Systems} (KGMS). A KGMS (Figure~\ref{fig:architecture}) performs complex rule-based reasoning tasks over very large amounts of data and provides methods and tools for data analytics and machine learning~\cite{BGPS17}.

\noindent

A KG can be viewed as a semi-structured data model
for the representation of domains of interest as facts, relationships, and attributes for those facts and relationships.
Facts and relationships can be either explicit, i.e., directly available in the enterprise data stores (relational databases, graph databases, NoSQL stores, RDF stores, OLAP systems, etc.)
or implicit, in the sense that they are generated when needed as the result of a reasoning task~\cite{BFGS19}.
Modern reasoning tasks embody both deductive and inductive aspects in that they
produce conclusions out of given axiomatic premises or ontological descriptions of the reality, but also
learn, enrich and improve these descriptions by means of empiric evidence in data, typical in pure machine learning approaches. 

\medskip
Technically, a KG can be defined as a semi-structured data model characterized by three components:
1.\ a \emph{ground extensional component}, with constructs (\emph{facts}) to represent
data in terms of a graph or a generalization thereof; 2.\ an \emph{intensional component}, with \emph{reasoning rules} over the
facts of the ground extensional component; 3.\ a \emph{derived extensional component},
produced in the \emph{reasoning process}, which applies
rules on ground facts~\cite{BeSG18}. 

\medskip
In this work, we refer to \emph{logic-based} KGs., i.e., where the intensional component is defined
with a logic-based KRR language.
Our reference KRR is \vadalog~\cite{BGPS17}, a language of 
the Datalog$^\pm$ family, which extends Datalog with existential quantification.
\vadalog strikes a good balance between expressive power and computation complexity.
It captures full \emph{Datalog}~\cite{datalog1,datalog3}, so it supports \emph{full recursion},
essential to navigate graph structures; at the same time it allows \emph{ontological reasoning},
being able to express SPARQL queries under set semantics and the entailment regime for OWL 2 QL
while guaranteeing \emph{scalability} thanks to \textsf{PTIME} data complexity for the reasoning task~\cite{GoPi15}.
A formal introduction to \vadalog is beyond the scope of this paper and can be found in~\cite{BeSG18}.

\medskip
Our perspective here is proposing KGs and reasoning methods and tools as
an ideal means to describe, conceptualize, and actively manage the highly interrelated data regarding
companies, with special reference to the crisis scenario of COVID-19. To this aim, specific models and problems
will be presented with a logic-based orientation and, when the case occurs, in the form of logical reasoning rules.
For this, we will resort to \vadalog syntax in such a way that
the meaning is intuitive, while for a thorough justification of the semantics, 
the reader is referred to specific literature~\cite{GoPi15}.

\section{The Knowledge Graph of Italian Companies}
\label{sec:italian_db}

In this section we consider the Italian company network before the COVID-19 outbreak and
provide a set of standard descriptive analytics. 

\begin{figure}[h!]
  \centering
  \includegraphics[width=1\textwidth]{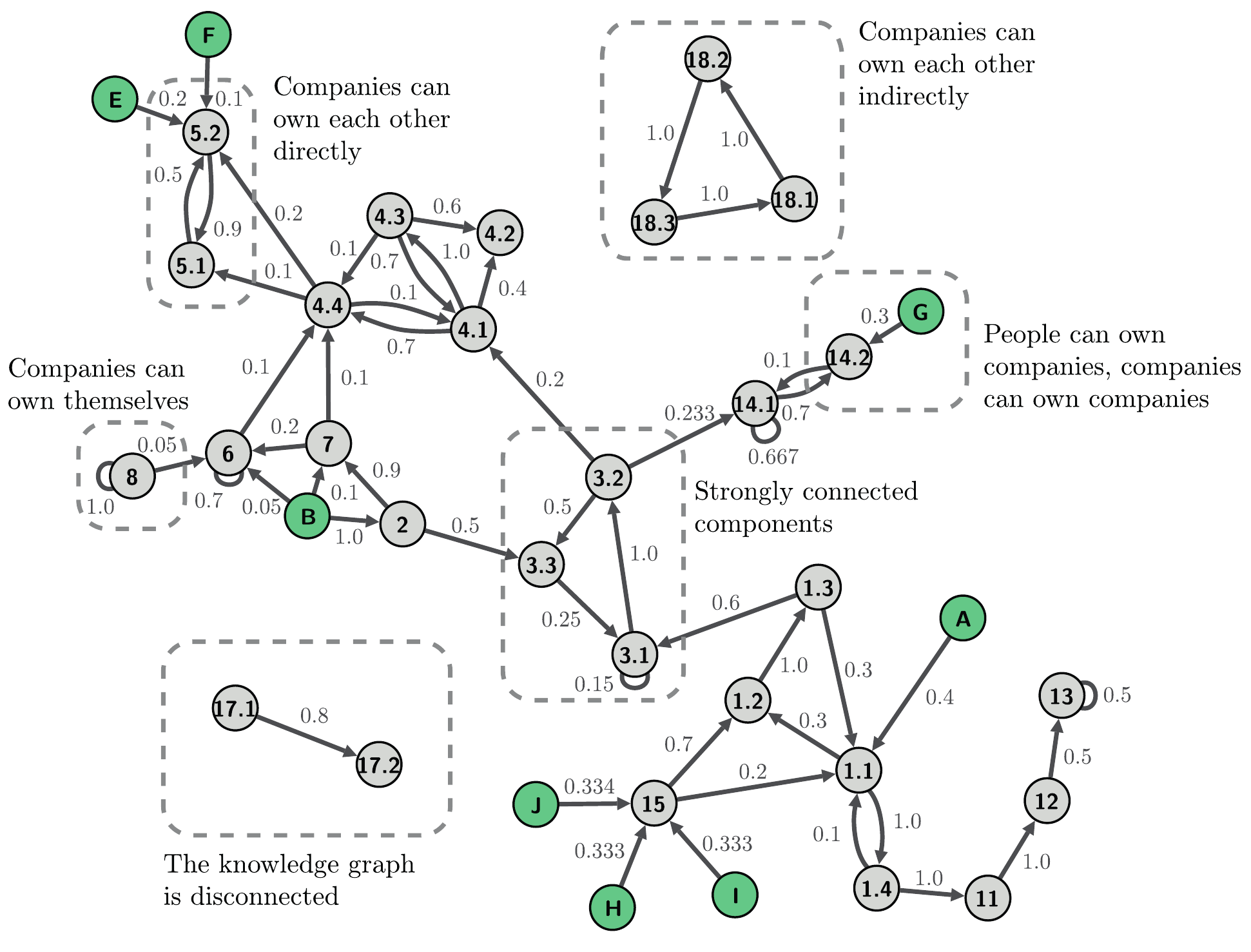}
  \caption{A simple company graph. Person nodes are labeled with letters A-J, companies with numbers. Numbers beginning with the same digits are in the same SCC (e.g., 1.1, 1.2). Ownership edges are directed from owner to owned.}.
  \label{fig:italian-graph-example}
\end{figure}

\medskip
In corporate economics, \emph{company ownership graphs} are essential objects~\cite{Barca01,Chapelle,Glat10,Rome15} playing an important role for central banks, financial authorities, and national statistical offices.
In such graphs, \emph{ownership} is the core concept: Nodes are companies and persons (grey nodes with numbers resp.\ 
green nodes with letters in Figure~\ref{fig:italian-graph-example}), and edges are ownership relationships labelled with the fraction of shares a company or person \(x\) owns of a company \(y\). Company graphs are helpful in many situations, for instance to compute the overall ownership a company retains of another, the control relationships, collusion phenomena, etc.

Central banks deal with data about companies to pursue core institutional goals in many relevant areas
such as \emph{banking supervision}, \emph{credit-worthiness evaluation}, \emph{anti-money laundering}, \emph{insurance fraud detection}, \emph{economic and statistical research}, and many more. 
In particular, the Bank of Italy owns the database of Italian companies provided by the Italian Chambers of Commerce. In recent work~\cite{ABIS20} we have motivated that many of the problems of interest on these data cannot be addressed
with standard techniques or database technology, but indeed require to be formulated as reasoning
tasks on KGs.

\medskip\noindent\textbf{The Italian Ownership Graph}. We consider the Knowledge Graph of Italian
companies as of 2018, containing the most updated data at our disposal.
We focus on \emph{non-listed companies}. 
For each of them, the graph contains several features including legal name, registered office address (geolocated), 
incorporation date, legal form, shareholders, and so on.
The data quality is high, though a marginal set of companies were
excluded from the KGs as attributes relevant for the analysis were missing or
it was impossible to determine the identity of the company.
A shareholder can be either a company or a person, with the standard anagraphic information.
Shares can be associated to different legal rights (e.g., ownership, bare ownership, etc.). 
We concentrate on all forms of ownerships and included in the KG
companies having at least one shareholder.

\begin{figure}[h!]
  \centering
  \includegraphics[width=1\textwidth]{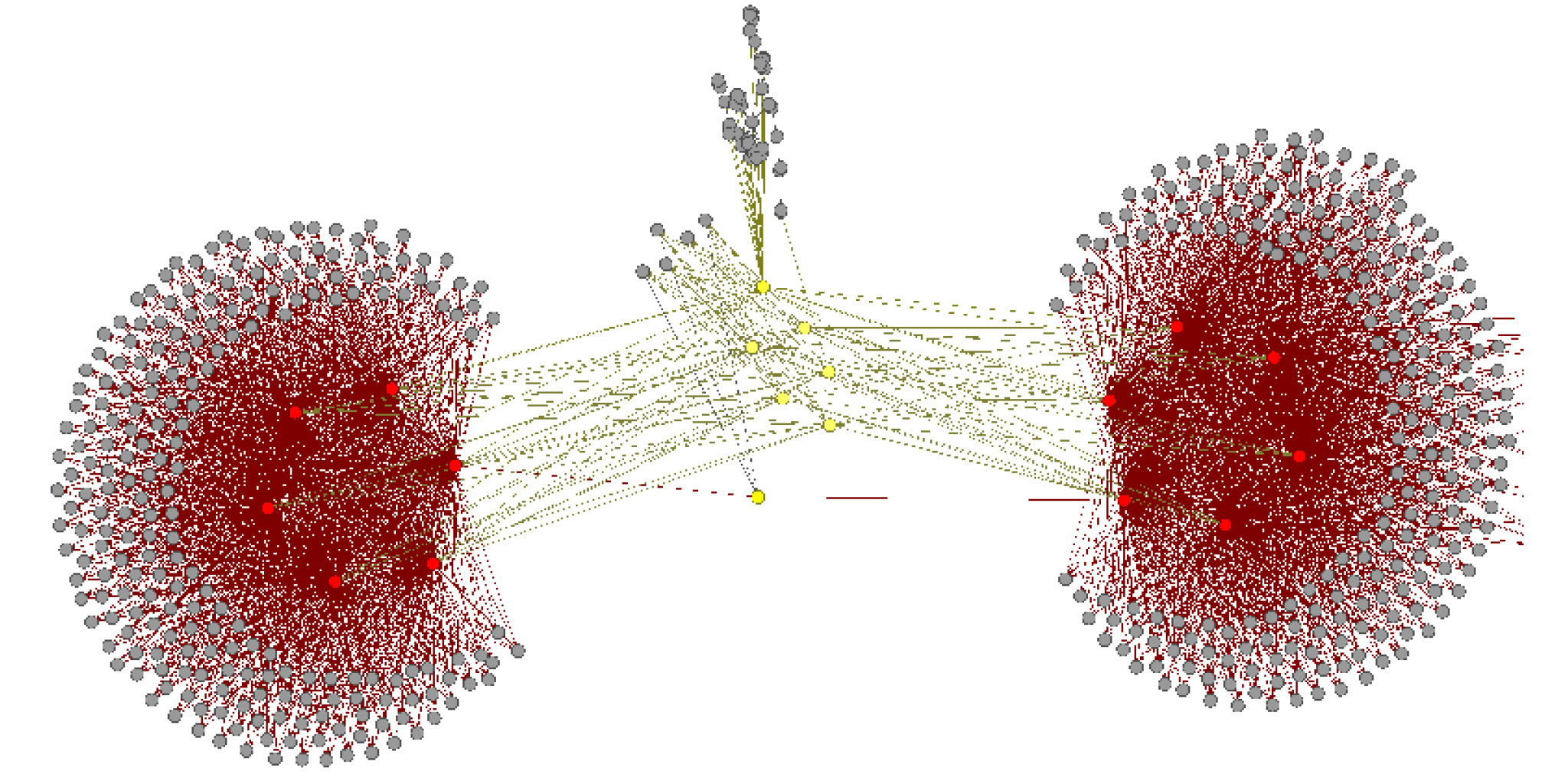}
  \caption{The ``lung graph''.}
  \label{fig:lungs-subgraph-example}
\end{figure}

\medskip
The graph counts 6.977M nodes (two thirds are people and one third are companies)
and 6.250M edges. It is quite fragmented and characterized by a very high number of Strongly Connected Components (SCC), 6.975M, 
often of one node, and 1.377M Weakly Connected Components (WCC), composed on average of 5 nodes. SCCs tend therefore to be small and, interestingly, the largest one has
only 13 nodes. Conversely, the largest WCC has more than 1.5M nodes. The \textit{average in-degree} is $\approx$ 2.43, the average \textit{out-degree} is $\approx$ 1.37 and the \textit{average clustering coefficient} is $\approx$~0.0059, very low with respect to the number of nodes and edges. Furthermore, it is interesting to observe that the \textit{maximum in-degree} is more than 3K and the maximum out-degree is more than 1K. The largest WCC represents a subgraph of the major Italian companies having the highest
out-degree. 
We also observe almost 3K self-loops, i.e., companies that 
subtract part of their own shares from the market~\cite{Glat10}.
The graph exhibits a \emph{scale-free network} structure~\cite{Bara99,HiBa08}: The degree distribution follows a power-law, with several 
nodes in the network acting as hubs.
An iconic representation of a portion of the company network, known as ``the lung graph'', originally appeared in~\cite{Rome15} and is shown in Figure~\ref{fig:lungs-subgraph-example}.

\section{Integrated Ownership and Company Control}
\label{sec:models}

Let us introduce some elements at the basis of our analytics on company ownership graphs, namely,
\emph{integrated ownership} and \emph{company control}. 
The former is a notion of accumulated ownership from a company $x$ to a company $y$: It accounts for
the ownership $x$ retains of $y$ along direct and all possible indirect connections. In terms of flow of
dividends, integrated ownership can be seen as the cumulative flow from $y$ to $x$, justified
by direct and indirect shareholding. On the other hand, company control represents whether $x$ can exert decision power on $y$.
As mentioned, these two concepts are independent, within certain limits.
We provide formal definitions for the above concepts as well as a specification of company control as a \vadalog reasoning task.

\medskip\noindent\textbf{Integrated ownership}. Let $G (N,E,w)$ be a directed weighted graph where $N=\{p_0,\ldots,p_n\}$ is a set of nodes, $E$ a set of edges of the form $(i,j)$, from node $i$ to node $j$, and $w : E \rightarrow \mathbb{R},~w\in (0,1]$ is a total weight function for edges. We denote by $w(p_i,p_j)$ the weight of edge $(i,j)$; an edge $(i,j)$ exists if and only if $w(p_i,p_j)\neq0$; furthermore self-loops are allowed, i.e. $i=j$.

A \emph{(directed) path} $P$ is a finite or infinite sequence $[p_1,\ldots,p_k]$ of nodes in $N$ such that $(i,i+1) \in E$ for every $i=1,\ldots,n$.  
We define the \emph{weight} $w(P)$ of a path $P$ as $w(P) = \Pi_{(p_i,p_j) \in P} w(p_i,p_j)$. 
For our purposes, nodes represent companies, edges $(i,j)$ represent ownership with share 

\begin{definition}\label{def:epsilon-baldone-path}
An $\epsilon-$Baldone path $P$ from $s$ to $t$ is a path $[s, p_1,\ldots, p_n, t]$ such that $s \neq p_i$ for $i=1,\ldots, n$
and $w(P)>\epsilon$, with $\epsilon \in \mathbb{R}^+$ and  $0 < \epsilon \leq 1$. We denote the weight of an $\epsilon-$Baldone path as $w_{\epsilon}(P)$.
\end{definition}

\noindent
Note that an \ebaldone path may have internal cycles 
(not involving $s$) or even be ultimately cyclic, with $s = t$.
Moreover, it can be infinite, in general. Intuitively, given two companies $s$ and $t$ we are interested in
knowing the amount of shares of $t$ owned by $s$. If we establish an acceptance threshold $\epsilon$,
then such ownership corresponds to the summation of the contribution of all the possibly infinite 
\ebaldone paths from $s$ to $t$, as follows.

\begin{definition}\label{def:epsilon-baldone-ownership-function}
\label{def:epsilon_baldone_ownership}
The $\epsilon$-\emph{Baldone ownership} of a company $s$ on a company $t$ in a graph $G$ is a 
function $\mathcal{O}_\epsilon^G(s,t): (N \times N) \rightarrow \mathbb{R}^+\cup\{\infty\}$ defined as
$(s,t) \rightarrow \sum_{P_i\in B_\epsilon}w_{\epsilon}(P_i)$, where $B_\epsilon$ is the set of all possible
$\epsilon$-Baldone paths from $s$ to $t$.
\end{definition}

\noindent
Although we have been referring to $s$ and $t$ as companies so far,
the discussion can be extended to people (i.e., physical shareholders) and companies, respectively, without loss of generality.
 We can now generalize \ebaldone ownership and consider all paths independently of their weight contribution by letting $\epsilon \to 0$.

\begin{definition}
\label{def:baldone_ownership}
The \emph{Baldone ownership} (which we will also refer to as \emph{integrated ownership})  of a company $s$ on a company $t$ in a graph $G$ is a 
function {$\mathcal{O}^G(s,t): (N \times N) \rightarrow \mathbb{R}\cup\{\infty\}$ defined as
$(s,t) \rightarrow \textit{lim}_{\epsilon \rightarrow 0}\mathcal{O}_\epsilon(s,t)$}.
\end{definition}

\noindent
Clearly, as a limit case of Definition~\ref{def:baldone_ownership}, for two non-mutually reachable nodes $x$ and $y$, 
we have that $\mathcal{O}^G(x,y)=0$. Moreover, observe that the convergence of integrated ownership is relevant
for its meaningfulness. We say that the integrated ownership of a company $s$ on a company $t$
over a graph $G$ converges if $\mathcal{O}^G(s,t) \le 1$; likewise, it converges for $G$
if $\mathcal{O}^G(s,t)$ converges for all $(s,t) \in E$. There are specific topological conditions, met by the graphs dealt with in this work that guarantee convergence and can be assessed via a complete graph traversal.
However, a thorough analysis of such conditions and the underlying motivations is beyond the scope of this paper and
can be found in~\cite{Glat10}.

\medskip
\noindent
Definitions~\ref{def:epsilon_baldone_ownership} and~\ref{def:baldone_ownership} can
be formalized as a \vadalog reasoning task, as follows.

\begin{align*}
  \textit{Own}(x,y,w), w>\epsilon, v = \textit{sum}(w), p=\textit{[x,y]} \to \textit{IOwn}(x,y,v,p). \tag{1} \\
  \textit{IOwn}(x,z,w_1,p_1),\textit{IOwn}(z,y,w_2,p_2),
  p = p_1 | p_2, \textit{BaldonePath}(p,v,\epsilon), \\v = \textit{sum}(w_1 \times w_2),
  \to \textit{IOwn}(x,y,v,p). \tag{2} \\
\end{align*}

\noindent
Whenever the direct ownership (atom \textit{Own}) a company $x$ 
has on the shares of a company $y$ is greater than a threshold $\epsilon$,
then $p$ is a valid \ebaldone path and $v$ is the
cumulative integrated ownership of $x$ on $y$ (Rule~(1)).
Integrated ownership from $x$ to $z$ can be composed with integrated
ownership from $z$ to $y$ if for the overall path $p$ 
(where the symbol ``$|$'' denotes the path concatenation operator)
from $x$ to $y$, Definition~\ref{def:epsilon-baldone-ownership-function} holds.
In such case, the integrated ownership is augmented with the $w_1 \times w_2$
contribution of path $p$ (Rule~(2)). Atom \textit{BaldonePath} represents whether
Definition~\ref{def:epsilon-baldone-path} holds.

\begin{figure*}[t!]
    \centering
    \begin{subfigure}[t]{0.45\textwidth}
        \centering
        \includegraphics[width=\textwidth]{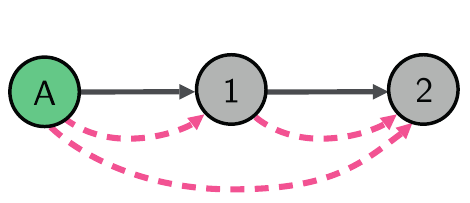}
        \caption{A simple indirect ownership. }
        \label{fig:essential-io-1}
    \end{subfigure}\hspace{0.05\textwidth}
    ~
    \begin{subfigure}[t]{0.45\textwidth}
        \centering
        \includegraphics[width=\textwidth]{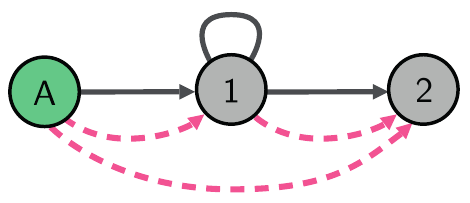}
        \caption{Indirect ownership with a self-loop.}
        \label{fig:essential-io-2}
    \end{subfigure}
    ~
    \begin{subfigure}[t]{0.7\textwidth}
        \centering
        \includegraphics[width=\textwidth]{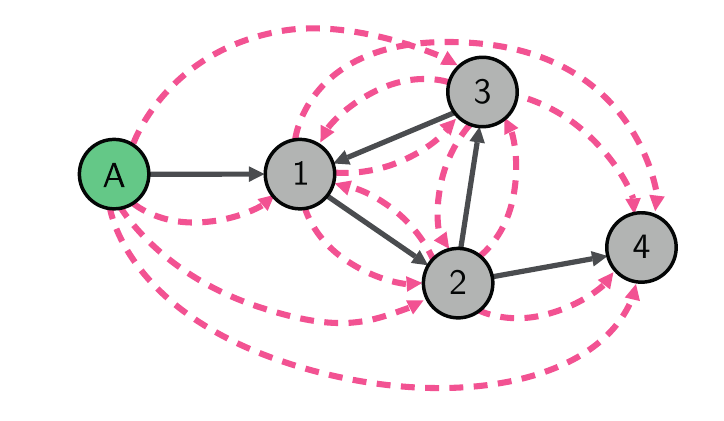}
        \caption{Indirect ownership with a strongly connected component.}
        \label{fig:essential-io-3}
    \end{subfigure}
    \caption{Cases of integrated ownership. Nodes with letters are people, nodes with numbers are companies, solid edges are direct ownership relationships
    while dashed edges represent integrated ownership.} 
    \label{fig:io-example2}
\end{figure*}

\medskip
Examples in Figure~\ref{fig:io-example2} show some basic cases of company ownership.
In Figure~\ref{fig:essential-io-1}, person $A$ has a non-null integrated ownership on company $2$ since she has a direct share of company $1$, 
and company $1$ has a direct share of company $2$. In Figure \ref{fig:essential-io-2}, the integrated ownerhship of person $A$ over company $1$ is more 
than the direct ownership along the edge $(A,1)$, since company $1$ has a partial ownership of itself, which in turn is owned by person $A$. 
Integrated ownership $(A,2)$ is increased as well, as it depends on $(A,1)$.
Finally, in Figure~\ref{fig:essential-io-3}, companies $1$, $2$ and $3$ form a SCC in which, by definition, all nodes own every other company in the component. 
So, if person $A$ owns a share of a company in a SCC, then she owns a share in all the companies of the SCC; 
if a company in the SCC owns a share in a company outside, then all companies of the SCC own a share in the same outside company.  

\begin{figure*}[t!]
    \centering
    \begin{subfigure}[t]{0.35\textwidth}
        \centering
        \includegraphics[width=\textwidth]{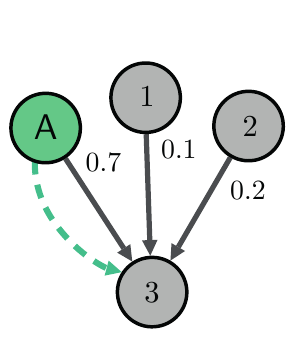}
        \caption{Direct control. }
        \label{fig:essential-control-1}
    \end{subfigure}\hspace{0.05\textwidth}
    ~
    \begin{subfigure}[t]{0.35\textwidth}
        \centering
        \includegraphics[width=\textwidth]{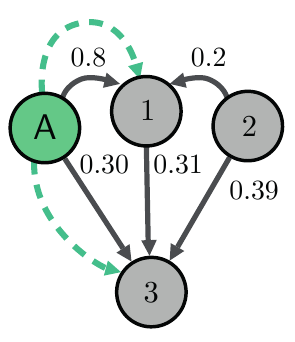}
        \caption{Indirect control.}
        \label{fig:essential-control-2}
    \end{subfigure}
\caption{Cases of company control. Nodes with letters are people, nodes with numbers are companies, solid edges are direct ownerships, dashed edges are control relationships.} 
\label{fig:io-example1}
\end{figure*}

\medskip\noindent\textbf{Company control}. Whilst ownership allows to measure mutual influence between companies,
company control is concerned with decision power, defining when a company  can induce decisions on another company.
A particularly clear formulation of company control has been provided in the logic/database literature~\cite{datalog1}. 

\begin{definition}
\label{def:control}
\noindent
A company (or a person) $x$ \emph{controls} 
a company $y$,  if: (i) $x$ directly owns more than $50\%$ of $y$; or, (ii)
$x$ controls a set of companies that jointly (i.e., summing the
shares), and possibly
together with $x$, own more than $50\%$ of $y$.
\end{definition}

\noindent
Definition~\ref{def:control} can be effectively formulated as a \vadalog reasoning task.
\begin{align*}
  \textit{Control}(x) & \to  \textit{Control}(x,x)    \tag{1} \\
  \textit{Control}(x,y), \textit{Own}(y, z, w), v = \textit{msum}(w, \langle y \rangle), v>0.5 &\rightarrow  \textit{Control}(x,z)   \tag{2} 
\end{align*}

\noindent
Given that every company has control on itself (Rule~(1)), we inductively
define control of $x$ on $z$, by forming the summation of shares of companies $y$ on $z$, 
over all companies $y$ controlled by $x$ (Rule~(2)).

\medskip\medskip
Figure~\ref{fig:io-example1} shows basic cases of company control.
In Figure \ref{fig:essential-control-1}, the first clause of Definition~\ref{def:control} applies: 
Person $A$ controls company $3$ since she has a direct share greater than 50\%.
In Figure \ref{fig:essential-control-2}, the second clause of Definition~\ref{def:control} applies: Person $A$ does not have enough directly owned share to control company $3$; 
nevertheless, she has an 80\% share on company $1$ and hence she controls it. Thus, the total share of company $3$ that $A$ controls 
rises to 61\%, which is the sum of the direct ownership $w(A,3) = 30\%$ and the ownership of controlled companies $w(1,3) = 31\%$. 
In the end, $A$ controls $3$.

\section{The ATECO Graphs}
\label{sec:ateco}

\begin{wrapfigure}{R}{0.5\textwidth}

\centering
\includegraphics[width=0.4\textwidth]{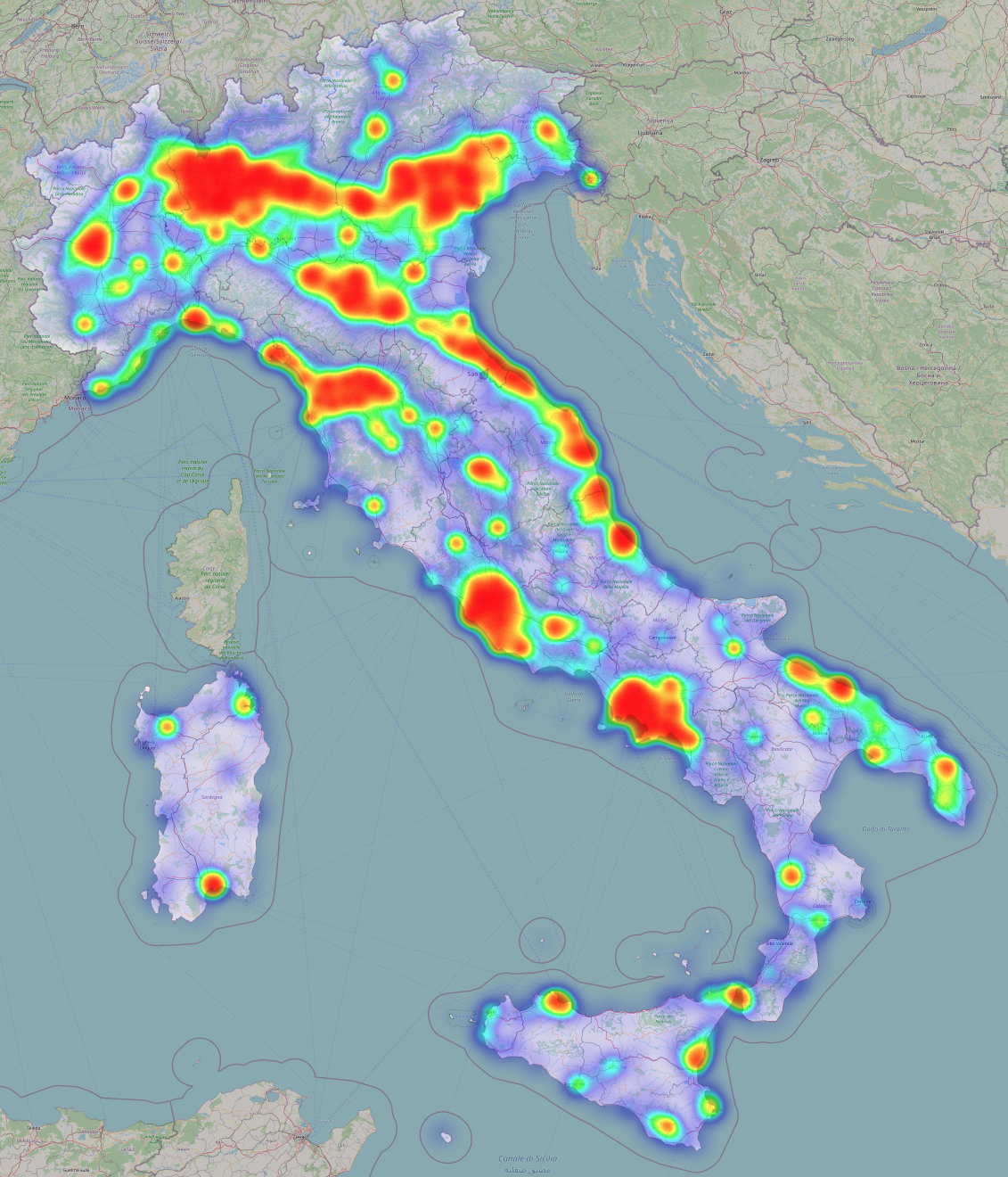}
\caption{Heatmap. Red resp. blue areas denote higher resp. lower impact of the
lockdown decrees on Italian companies.}
\vspace{-7mm}
\label{fig:heatmaps}
\end{wrapfigure}

In this section we provide basic descriptive analytics of the KGs built with various configurations,
during the COVID-19 outbreak in Italy, defining which economic activities (identified via the ATECO code) are
allowed or forbidden, as considered strategic or essential at a specific point in time. 
More detailed analytics, considering company groups (namely, conglomerates)
are dealt with in Section~\ref{sec:conglomerates}. 
The datasets for specific KGs are made available upon request.

\medskip\noindent\textbf{Configurations}.
We built different subgraphs of the Italian ownership KG  described in Section~\ref{sec:italian_db}, with 
the following configurations:  \emph{ATECO3.22}, after the issuance of Italian decree
22.03.2020~\cite{Decree22032020} defining preliminary shutdown of non-strategic economic activities; 
\emph{ATECO3.25}, after the issuance of Italian Decree 25.03.2020~\cite{Decree25032020} clarifying the former and
modifying the shut-down activities; \emph{ATECO4.10}, after the issuance of Italian Decree 10.04.2020~\cite{Decree10042020},
with partial re-openings. We also considered the special provisions of Italian regions, allowing or forbidding
specific economic activities (e.g., Lombardia, Veneto, etc.).
So far, local (city-level) provisions and ad-hoc exceptions have not been considered.

\medskip
Each KG $G$ has been built as a subset of the general Italian ownership graph 
s.t.: A company is in $G$ if its ATECO is allowed by the respective decree; a person is in $G$ if
she owns a percentage of a company in $G$; ownership edges are in $G$ if they connect two
companies (or a person and a company) both in $G$. 
Let us see the main characteristics of each subgraph.

\medskip\noindent\textbf{ATECO3.22}.
The graph has 2.141M nodes, with 1.345M  person nodes and 796K  company nodes, 
and 1.660M edges. The graph inherits the same kind of fragmentation as the original structure described in Section~\ref{sec:italian_db}. Indeed, it has a high number of SCCs, 2.140M, with a maximum size of 8 nodes. The number of WCCs is 570K with an average size of $\approx$ 4 nodes and a maximum size of 217K.

\medskip\noindent\textbf{ATECO3.25}.
The graph has 2.156M nodes, where 1.351M are person nodes and 805K are company nodes, 
and 1.670M edges.  The number of SCCs is slightly increased with respect to ATECO3.22, 2.155M, with a maximum size of 8 nodes. The number of WCCs is 570K with an average size of $\approx$ 4 nodes and a maximum size of 215K nodes.

\medskip\noindent\textbf{ATECO4.10}.
The total number of nodes is 2.225M, with 1.395M person nodes and 830K company nodes. The graph presents a large number of SCCs, 2.225M, with an average of 1 node per SCC and a maximum of 8 nodes in the largest SCC. There are 595K WCCs, where the largest is composed by 228K nodes and the average size is $\approx$ 4. 

\medskip All the graphs present \textit{in-degree} and \textit{out-degree} average values of $\approx$ 2.10 and $\approx$ 1.19 respectively. These values are consistent with the ones obtained for the original graph of Section~\ref{sec:italian_db}.

\medskip The subgraphs generated for the decrees issued so far show small differences in terms
of aggregate analytics, an expected result provided that the initial
Italian Decree 2020.03.25~\cite{Decree25032020} applied very restrictive measures
and resolutions relaxing such restrictions have not been issued yet.
Nevertheless, monitoring the evolution of standard network analytics is matter of ongoing
work, which will get more and more significant as the situation evolves and is 
at the basis of the more advanced analyses dealt with in the next sections.

\medskip
The heatmap in Figure~\ref{fig:heatmaps} shows the impact of 
the first Italian decree on the Italian companies, highlighting the
areas with higher density of closed companies.

\section{Evaluating the Conglomerates}
\label{sec:conglomerates}

While there are no agreed definitions of \emph{company group} or \emph{company conglomerate},
the generally intended meaning is that of a corporate group composed of multiple business entities
operating together for some purpose, including: Market differentiation, reuse of existing
plant facilities, increasing the offer with a wider range of products, risk differentiation and mitigation~\cite{brit}.
A conglomerate often involves large national or multinational groups and
various corporate patterns, e.g., with one parent company, controlling the others, and many subsidiaries.

\medskip
The organization of companies in conglomerates is an important factor for their resilience, as
differentiating by economic activity or geography can help control the effects induced by
production restrictions such as the Italian limiting decrees discussed in Section~\ref{sec:ateco}.

\medskip In this section we first propose a model to identify conglomerates
in Company Ownership Graphs, based on the notion of integrated ownership (Definition~\ref{def:baldone_ownership}) introduced in Section~\ref{sec:models}.
Then, we analyze again the configurations of the Italian graph illustrated in Section~\ref{sec:ateco} under the
conglomerate perspective and contribute a set of more advanced analytics.

\medskip
The intuition behind our notion of conglomerates is that each company acts as an ``attractor'',
in the sense that given that a certain company $x$ belongs to a conglomerate $C$,
then any company $y$ that is sufficiently close to $x$, will be part of $C$ as well.
Thus, our model is not based on the classic concept of cluster, where all the elements
are within a certain distance from a centroid, but, instead, a conglomerate is a \emph{topological space}
induced by a binary relation representing \emph{ownership vicinity} between neighbouring companies.
It turns out that a suitable distance can be formulated by generalizing the concept of \ebaldone ownership  (Definition~\ref{def:epsilon_baldone_ownership})
making it symmetric, as follows.

\begin{definition}\label{def:undirected_epsiolon_baldone_ownership} The \emph{undirected \ebaldone ownership} of a company $s$ on a company $t$
in a graph $G$ is a function $\mathcal{U}^G_\epsilon(s,t) : \text{max} \{O^G_\epsilon(s,t), O^G_\epsilon(t,s)\}$.
\end{definition}

\noindent
We can now define our concept of vicinity, key to forming conglomerates.

\begin{definition}\label{def:same-conglomerate-c-close}
The binary relation \emph{vicinity} $\mathcal{V}_\epsilon^G$ holds between two companies $x$ and $y$ of a graph $G$ (and we say --- they are close ---), if:
(i) $\mathcal{U}^G_\epsilon(x,y) > 0$; or, (ii) there exists a third party $z$ of $G$ such that
$\mathcal{U}^G_\epsilon(z,x) > 0$ and $\mathcal{U}^G_\epsilon(z,y) > 0$.
\end{definition}

\noindent
Interestingly, not only are two companies close when there is a non-null \ebaldone path
between them (independently of the orientation), but it may also be the case that $x$ and $y$ are not
mutually reachable and yet close. For this to hold, we need a third-party sufficiently close to both.
Although we have provided a definition on companies, the third party can actually be a
physical shareholder as well, and we actually considered these cases.
Vicinity is parametrized in $\epsilon$, as it depends on the tolerance we assume.
Observe that as a limit case, if we define vicinity on Baldone ownership, so for $\epsilon \rightarrow 0$,
then it degenerates into reachability of company nodes over $G$ (assuming the absence of null edges).
We are now ready to define conglomerates.

\begin{definition}\label{def:conglomerate}
An \econglomerate is an equivalence class of $\mathcal{V}_\epsilon^{+G}$, where $+$ denotes
the transitive closure of the vicinity relation defined on a graph $G$.
\end{definition}

With the above definition, the conglomerates of an ownership graph
can be modeled as a reasoning task in \vadalog as follows:

\begin{align*}
  \textit{O}(a,b,x),x>\epsilon \to \textit{U}(a,b,x),\textit{U}(b,a,x) \tag{1} \\
  \textit{U}(a,b,x), a>b \to \exists z~\textit{C}(z,a),\textit{C}(z,b)  \tag{2}  \\
  \textit{U}(w,a,x), \textit{U}(w,b,x) \to \exists z~\textit{C}(z,a),\textit{C}(z,b)  \tag{3} \\
  \textit{Company}(a) \to \exists z~\textit{C}(z,a) \tag{4} \\
  \textit{C}(y,a), \textit{C}(x,a) \to x=y  \tag{5} \\
\end{align*}

\noindent
Atom $O$ in Rule~(1) denotes Baldone ownership as in Definition~\ref{def:baldone_ownership}.
It is then ``copied'' into $U$, which denotes the undirected \ebaldone ownership as in Definition~\ref{def:undirected_epsiolon_baldone_ownership},
by filtering on the appropriate value for $\epsilon$. In particular, $\epsilon$ represents our
\emph{conglomerate threshold}, that is, the minimum ownership value
required to extend by one company the topological space of a conglomerate.
Rule~(2) is then the basic extension rule, asserting that whenever two companies $x$ and $y$ are close enough
(more than $\epsilon$), there exists a conglomerate $z$ to which both belong.
The third-party case is captured by Rule~(3), where the existence of a company $z$ close
to $a$ and $b$ implies that they belong to the same conglomerate $z$. Rules~(4-5) are
consistency rules, to guarantee that each company belongs to exactly one conglomerate.

\begin{figure}[h!]
  \centering
  \includegraphics[width=0.7\textwidth]{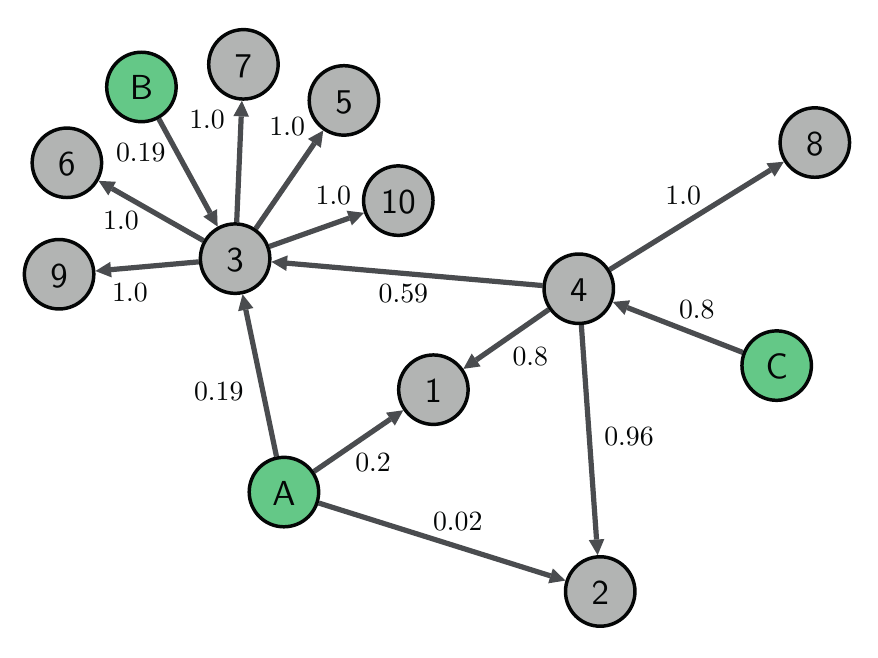}
  \caption{Sample anonymized conglomerate from the Italian graph.}
  \label{fig:conglomerates-example}
\end{figure}

\noindent

Figure~\ref{fig:conglomerates-example} shows a sample $\epsilon$-conglomerate
extracted from the Italian graph by posing $\epsilon > 0.5$.
Such threshold captures a sort of share majority: A company within 
a certain conglomerate acts as an attractor for another company
if it holds the majority of the shares (or vice versa, given that
$\mathcal{V}_\epsilon^G$ is symmetric). The lower the threshold,
the lower the conglomerate cohesion will be, with larger conglomerates
including many loosely connected companies.
Conversely, the higher the
threshold, the more cohesive the conglomerate will be and involve
companies with higher mutual integrated ownership.
In the figure, edges with share greater than 0.5 show cases where two nodes are close
(Definition~\ref{def:same-conglomerate-c-close}(i)), such as 3-9 and 3-5;
pairs of nodes whose share majority is held by 
the same third party are close as well 
(Definition~\ref{def:same-conglomerate-c-close}(ii)), for instance 
companies 9 and 5 are both fully owned by 3 and so are close.
Therefore, 3, 5 and 9 are all in the same equivalence class of  $\mathcal{V}_\epsilon^{+G}$, hence in the same conglomerate.

\medskip 
\medskip\noindent\textbf{Conglomerates in the Italian Ownership Graph}. We applied Definition~\ref{def:conglomerate} to the different versions of the Italian KG:
the baseline graph before the COVID-19 outbreak (described in Section~\ref{sec:italian_db})
and the ATECO KGs, capturing the company network at the various pandemic
phases in correspondence to the issuance of Government decrees (described in Section~\ref{sec:ateco}).
In particular, we decomposed each graph into its constituent conglomerates. 
For this, we set $\epsilon=0.5$ in order to
identify conglomerates where there is a strong majority influence
between the involved companies (i.e., control is likely to hold) and
excluded trivial conglomerates with just one company.
We computed a set of basic descriptive indicators,
reported in Figure~\ref{fig:conglomerates-stats}.
\footnote{The computation of the conglomerates 
in this table was carried out using the \vadalog 
engine and the set of rules is made available upon request; 
the computation took $\approx 36$ million deduction steps 
and was completed in $\approx 2$ hours on a modern cloud machine (8 cores, 64 GB RAM). No other system in use at our company could have matched this performance or even fully expressed the problems tackled here.} 
The exhaustive breakdown of all the company groups is made available upon request.

\begin{figure}
\begin{center}
 \begin{tabular}{|p{.13\textwidth}||c|c|c|c|c|c|}
 \hline
  {\scriptsize $\epsilon=0.5$} & {\tiny \textbf{Conglomerates}} & {\tiny \textbf{\begin{tabular}[c]{@{}c@{}}Companies per \\ cong.\end{tabular}}} & {\tiny \textbf{\begin{tabular}[c]{@{}c@{}}ATECO per \\ cong.\end{tabular}}} & {\tiny \textbf{\begin{tabular}[c]{@{}c@{}}Max \\ cong. size\end{tabular}}} & {\tiny \textbf{\begin{tabular}[c]{@{}c@{}}Max ATECO\\ per cong.\end{tabular}}} & {\tiny \textbf{\begin{tabular}[c]{@{}c@{}}Regions per \\ cong.\end{tabular}}}\\
 \hline
 \hline
{\tiny \textbf{Baseline}} & \scriptsize 283281&\scriptsize 2.82& \scriptsize 2.28&\scriptsize 21436&\scriptsize 1041&\scriptsize 1.16\\
 \hline
 {\tiny \textbf{ATECO3.22~}}~ &   \scriptsize 55490  & \scriptsize 2.56   &\scriptsize 1.99& \scriptsize 2269& \scriptsize 336& \scriptsize 1.17\\
 \hline
 {\tiny \textbf{ATECO3.25~}} &\scriptsize 56441 & \scriptsize 2.57&  \scriptsize 2.00 & \scriptsize 2579& \scriptsize 349& \scriptsize 1.17\\
 \hline
 {\tiny \textbf{ATECO4.10~}} &\scriptsize 58265 & \scriptsize 2.57&  \scriptsize 2.00& \scriptsize 2746& \scriptsize 369& \scriptsize 1.17\\
  \hline
 \end{tabular}
\end{center}
\medskip
\caption{A set of basic descriptive indicators for company conglomerates.}
\label{fig:conglomerates-stats}
\end{figure}

An exhaustive analysis of the data in Figure~\ref{fig:conglomerates-stats} is
beyond the scope of this work and matter of follow-up economic research. 
It must be anyway noticed that 
the number of conglomerates drops with the first restrictive decrees,
which have a disgregating effect on the company network.
Surprisingly, big conglomerates do not survive the lockdown
unlike average-sized ones. From another perspective, it looks like the few
very big Italian conglomerates are not resilient enough, while
the smaller ones uniformly absorb the effects of production limitations.
The regional distribution appears very limited in Italy or anyway
insufficient to control the breakdown of company groups in
a global scenario where, while some regions have been particularly
affected, the emergency and the regulations interested the whole country.
It will be interesting, on the other hand, to assess the evolution of the indicators with the gradual and possibly local relaxation of the regulations.

\section{Reasoning on Golden Power}

\label{sec:golden_power}

While our focus in the previous sections was on Knowledge Graph construction and analytics, in this section we are going one step further and show an example of KG-based applications  that allow to 
support policy
decisions to be made, advice to be given to companies, as well as proactive actions to be taken. More specifically, we present a number of KG-based reasoning 
tasks that revolve around the concept of \emph{Golden Power}, which we will introduce next.

During crises, share prices of most industries typically get lower and, as not all countries face the crises (or not all at the same time 
or of the same magnitude), 
foreign takeovers are possible. Italy has a powerful legal tool, called ``Golden Power''~\cite{GovernmentGoldenPower,GrecoGoldenPower}, that allows the Government to veto specific transactions that would cause a change of the control for strategic firms.\footnote{The 
concept of ``strategic'' firm/company is somehow overloaded in the literature
and interesting definitions have been proposed in different contexts~\cite{BeMa08,CaEG97}.
Here we simply refer to a company of national relevance in the broadest sense.} 
A well-known example of this law in action
was in 2018, with the fiber-optic network group Retelit~\cite{RetelitGoldenPower}. Now, as Italy is one of the most hardly stricken countries by the COVID19, the government is ready to enforce this law~\cite{PetriniGoldenPower,reuters1}. In this work, we give formulations of KG-based reasoning tasks, about Golden Power that provide insights on:

\begin{enumerate}
\item \emph{detecting} cases of transactions hiding possible takeover attempts;
\item \emph{suggesting} the limits within which Golden Power may be exercised;
\item giving options for \emph{proactively protecting} companies from takeover attempts.
\end{enumerate}

\noindent
After these three central problems, we consider two other advanced cases, namely, how to deal with possible \emph{collusion} between companies, and how to deal with \emph{missing or incomplete} information.

Finally, we recall that while 
we focus on Italy here
(for which we can offer a concrete insight into the specific Golden Powers), similar frameworks exist in many jurisdictions, and we believe that general insight into the analysis of Golden Power-like problems can be obtained from our findings.

\begin{figure}[t]
\begin{mdframed}[frametitle={%
        \tikz[baseline=(current bounding box.east),outer sep=0pt]
        \node[anchor=east,rectangle,fill=blue!20]
        {\strut Golden Power Check};},linecolor=blue!20,linewidth=2pt,frametitleaboveskip=\dimexpr-\ht\strutbox\relax]

\begin{tabular}{l p{9.25cm}}
\emph{Goal}: & The general goal is checking whether an acquisition (of shares, stocks, etc.) causes any strategic Italian company to 
become
controlled by a foreign company.  \\[.4em]
\emph{Setting}: & Let $S$ be a set of strategic companies and $F$ be a set of foreign companies. Let $t$ be a transaction (e.g., an offer issued by a company $x$ to buy an amount $s$ of shares of a company $y$), with $x,y \in S \cup F$. \\[.4em]
\emph{Question}: \protect{   } & Decide whether $t$ causes any company in $F$ to control a company in $S$. \\[.4em]
\emph{Insight}: \protect{   } & Consider exerting ``golden power'' to block $t$. \\
\end{tabular}
\end{mdframed}

\vspace*{.5em}

\centering
  \includegraphics[width=0.75\textwidth]{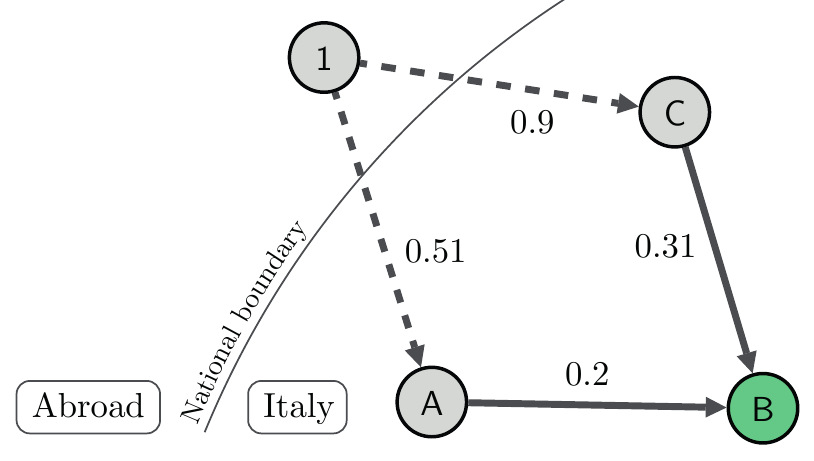}

\caption{Definition (top) and example (bottom) for Golden Power Check.}
\label{fig:goldenpowercheck}
\end{figure}

\medskip
\noindent
\textbf{Golden Power Check}.
The first, most direct problem is the detection of a transaction where a strategic Italian company would be taken over, and hence where golden power may be an option to take into account. We call this problem \emph{Golden Power Check}. The definition and an example are given in Figure~\ref{fig:goldenpowercheck}.

\smallskip
\noindent
\emph{Example}.
Let us consider the example shown in the bottom part of Figure~\ref{fig:goldenpowercheck} in detail. We first consider the setting. Circular nodes represent companies, i.e., companies $1$, $A$, $B$, and $C$. The national boundary of Italy is shown, with company $1$ outside this national border\footnote{A company incorporated or organized under the laws of 
another country.} and companies $A$, $B$, and $C$ inside this national border. In our case, company $1$ is in the set of foreign companies under investigation (forming the set $F$ in the definition shown in Figure~\ref{fig:goldenpowercheck}). The differently colored node $B$ is in the set of strategic companies (forming the set $S$ in our definition shown in Figure~\ref{fig:goldenpowercheck}).
Solid lines denote ownership, e.g., company $C$ owns $31\%$ of $B$.

Candidate transactions are shown via the dashed edges. Our first candidate transaction is $t_1$ where an ownership of $51\%$ of $A$ is acquired by $1$. The second candidate transaction is $t_2$ where an ownership of $90\%$ of $C$ is acquired by $B$. Let us first consider transaction $t_1$ as our transaction of interest. This would give $1$ control of $A$, and hence a $20\%$ ownership of $B$. So far, the total ownership of strategic company $B$ by company $1$ is thus $20\%$. This is fine, and there is no need to block transaction $t_1$.

Now assume that transaction $t_1$ was processed (i.e., it becomes a solid line). Now we consider transaction $t_2$ where an ownership of $90\%$ of $C$ is obtained by $1$. This would give $1$ control of $C$ and hence $31\%$ ownership of $B$. Together with the ownership of $20\%$ of strategic company $B$ that $1$ already holds, it now has $51\%$ ownership of company $B$ and thus controls it. Transaction $t_2$ must be blocked using Golden Power if the strategic company $B$ should not come under the control of $1$. 
Finally, we remark that if transaction $t_2$ would have come before $t_1$, it would have been fine to process $t_2$ and block $t_1$. This concludes our example.

This was certainly a very simple example, as all ownership was via at most one intermediate company. In reality, we know that (i) ownership is often obtained via many intermediate companies and (ii) much more complex settings such as cycles occur. This adds quite a lot complexity to this reasoning task.

The definition of Figure~\ref{fig:goldenpowercheck} can be effectively formulated as the following \vadalog reasoning task:
\begin{align*}
  \textit{T}(x,y,w) & \to \textit{Own}(x,y,w) \tag{1} \\
  \textit{Control}(x,y) & \to  \textit{Control}(x,x)    \tag{2} \\
  \textit{Control}(x,y), \textit{Own}(y, z, w), v = \textit{msum}(w, \langle y \rangle), v>0.5 &\rightarrow  \textit{Control}(x,z)   \tag{3} \\
  \textit{F}(x), \textit{S}(y), \textit{Control}(x,y) & \to \textit{GPCheck}(x,y) \tag {4}
\end{align*}

\noindent
Rule 2 and 3 define control based on ownership, as we have seen already in Section~\ref{sec:models}. Rule 1 defines that for the purpose of our analysis, we want transaction $t$ to be considered applied, i.e., leading to ownership. Finally, Rule 4 formulates our question, namely computing all companies in $F$ that control at least one company in $S$. If \textit{GPCheck} is empty, there is no reason for Golden Power to be exerted; in case it is non-empty it will give a table of the companies which witness the takeover.

\begin{figure}[t]
\begin{mdframed}[frametitle={%
        \tikz[baseline=(current bounding box.east),outer sep=0pt]
        \node[anchor=east,rectangle,fill=blue!20]
        {\strut Golden Power Limit};},linecolor=blue!20,linewidth=2pt,frametitleaboveskip=\dimexpr-\ht\strutbox\relax]

\begin{tabular}{l p{9.25cm}}
\emph{Goal}: & The goal is computing the maximum amount of share a company $z$ can buy of a company $x$ without controlling any company in a set $S$.  \\[.4em]
\emph{Setting}: & Let $S$ be a set of strategic companies and $F$ be a set of foreign companies. Let $t$ be a transaction (e.g., an offer issued by a company $x$ to buy an amount $s$ of shares of a company $y$), with $x,y \in S \cup F$. \\[.4em]
\emph{Question}: \protect{   } & Find the maximum value for $s$ s.t.\ there are no companies in $F$ that control companies in $S$.
 \\[.4em]
\emph{Insight}: \protect{   } & Transactions that acquire (much) less
than $s$ shares of $y$, mean $y$ does not need
to summon Golden Powers.\\
\end{tabular}
\end{mdframed}

  \centering
  \includegraphics[width=0.75\textwidth]{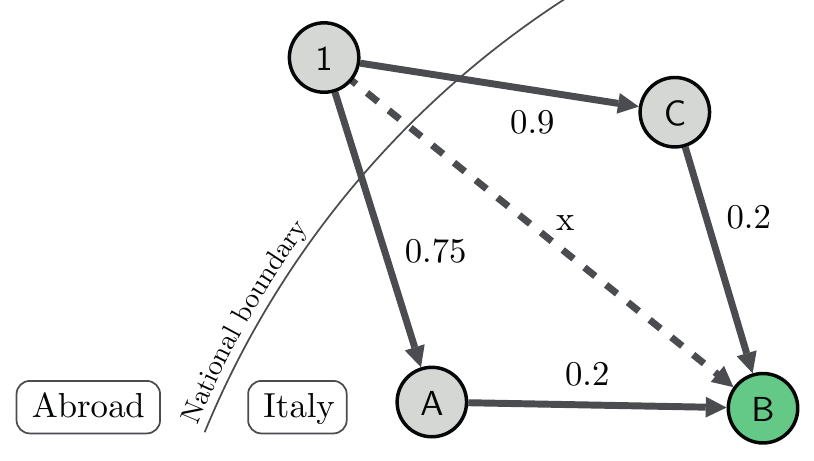}

\caption{Definition (top) and example (bottom) for Golden Power Limit.}
\label{fig:goldenpowerlimit}
\end{figure}

\medskip
\noindent
\textbf{Golden Power Limit}.
The second relevant problem is about \emph{advising} companies regarding the limits transactions may be allowed to take place (without requiring the use of golden power to prevent a takeover attempt). We call this problem \emph{Golden Power Limit}. The definition and an example are given in Figure~\ref{fig:goldenpowerlimit}.

\medskip
\noindent
\emph{Example}.
Let us consider the example shown in the bottom part of Figure~\ref{fig:goldenpowerlimit} in detail. We see a setup similar to the example described in Figure~\ref{fig:goldenpowercheck}, with the same companies, but with different ownership edges. In our example we see company $1$ owning $75\%$ of $A$, and in turn, $A$ owning $20\%$ of strategic company $B$. This gives $1$ a $20\%$ ownership of $B$. Similarly, company $1$ has a $90\%$ ownership of $C$, and in turn, $C$ owns $20\%$ of $B$. This gives $1$ another $20\%$ ownership of $B$, coming up to a total of $40\%$. 

The question we ask here -- which is very easy to answer in this simple example -- is for a proposed transaction to determine what is the maximum amount $x$ of ownership that can be obtained without obtaining control resp.\ without having Golden Power to be exercised.
In our case $t$ is the transaction of $1$ buying $x$ ownership of $B$. It is in this simple case clear that $x = 10\%$ as anything higher would give $1$ control over $B$. This concludes our example.

Again, this was a very simple example and as noted before as points (i) and (ii), indirect ownerships and cyclic structures may make this reasoning problem much more complex. Also, we considered as $t$ a direct transaction from the company under investigation to the strategic company. In fact, it could be (iii) a transaction between two national companies (both within Italy) that needs to be investigated.

\begin{figure}[t!]
\begin{mdframed}[frametitle={%
        \tikz[baseline=(current bounding box.east),outer sep=0pt]
        \node[anchor=east,rectangle,fill=blue!20]
        {\strut Golden Power Protection};},linecolor=blue!20,linewidth=2pt,frametitleaboveskip=\dimexpr-\ht\strutbox\relax]

\begin{tabular}{l p{9.25cm}}
\emph{Goal}: & The goal is computing the share increment needed by publicly controlled companies to prevent foreign takeovers.  \\[.4em]
\emph{Setting}: & Let $S$ be a set of strategic companies and $F$ be a set of foreign companies.  Let $P$ be a set of publicly controlled companies (s.t.\ $P$ is disjoint from $F$ and $S$). \\[.4em]
\emph{Question}: \protect{   } & Determine the acquisitions of shares needed from companies in $P$ to companies in $S$ such that it is impossible to issue any set of transactions $t$, where $t$ is a transaction from $x$ to $y$, with $x,y \in S \cup F$ such that a company in $F$ controls one in $S$.
\\[.4em]
\emph{Insight}: \protect{   } & Consider buying (temporarily) shares of $S$ via $P$ to prevent takeovers without resorting to Golden Powers.
 \\
\end{tabular}
\end{mdframed}

  \centering
  \includegraphics[width=1\textwidth]{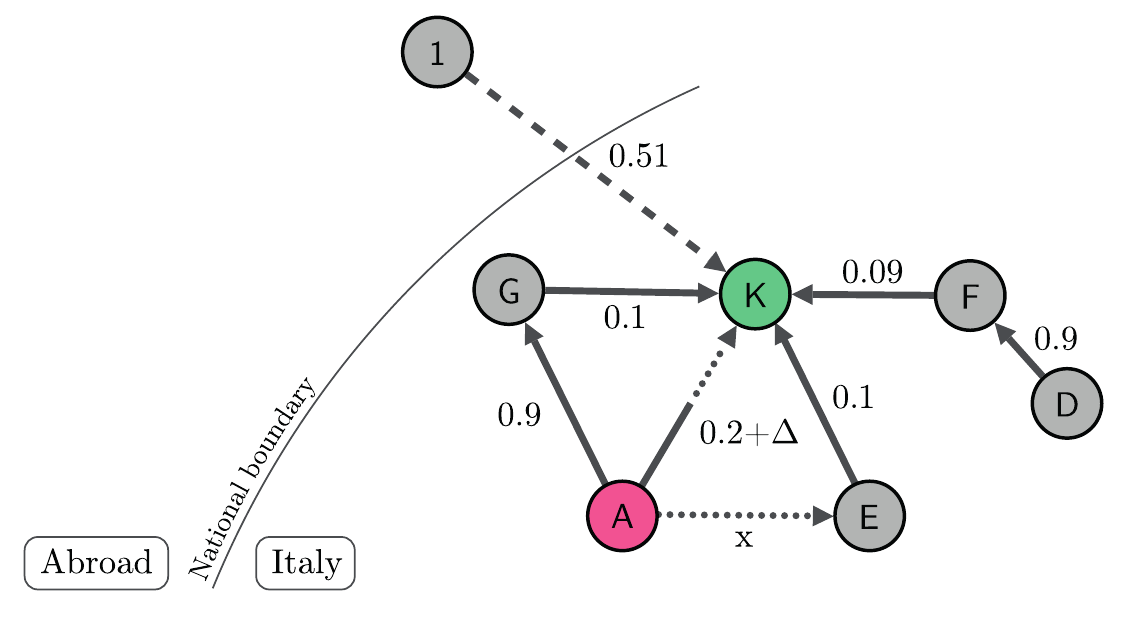}

\caption{Definition (top) and example (bottom) for Golden Power Protection.}
\label{fig:goldenpowerprotection}
\end{figure}

\medskip
\noindent
\textbf{Golden Power Protection}.
The third basic problem is for \emph{proactively preventing} that the use of golden power becomes necessary. The use of golden power comes with political and economic consequences, so in many cases it may be desirable to proactively prevent takeovers. We call this problem \emph{Golden Power Protection}. The definition and an example are given in Figure~\ref{fig:goldenpowerprotection}.

\medskip
\noindent
\emph{Example}.
Let us consider the example shown in the bottom part of Figure~\ref{fig:goldenpowerprotection} in detail. Marked in this example we see a number of different types of categories. Company $K$ is our strategic company, and again company $1$ is a foreign company. In addition, we have company $A$ as a publicly controlled company (that is, in the set $P$ of our definition). In this simple example, the question of how to protect $K$ from takeover attempts (such as the transaction of $1$ obtaining $51\%$ ownership of $K$ shown via the dashed lines) is reasonably simple.

As we have only one publicly controlled company $A$ in our case, and only one strategic company $K$, it is clear that we need to ``beef up'' the already existing ownership of $20\%$ that $A$ has on $K$ by some $\Delta$. The answer in our case is that $\Delta = 21\%$: $A$ directly owns $20\%$ of $K$ and indirectly via $G$ owns another $10\%$ of $K$, totaling $30\%$. Thus, with another $21\%$, takeover attempts are prevented.

We also consider a variant of this scenario with we call \emph{Golden Power Protection with Intermediaries} (and in this case call the standard variant \emph{Golden Power Protection without Intermediaries}). Also observe that the definition given in Figure~\ref{fig:goldenpowerprotection} only allows publicly controlled companies to obtain any ownership of $K$. If we relax the definition to publicly controlled companies being allowed to obtain any ownership (not necessarily of $K$, i.e., with \emph{intermediaries}) we see another scenario possible:

Company $A$ could obtain an amount $x$ of $E$, together with an amount $\Delta$ of $K$. In our case, $x=51\%$ and the remaining $\Delta=11\%$ would be sufficient. This may help especially in situations where company $E$ owns parts of other companies apart from $K$, or in case ownership of $21\%$ is unrealistic to be directly obtained of $K$. This concludes our example.

\medskip
\noindent
\textbf{Advanced cases}.
Beyond the cases discussed so far, there are manifold variations possible of the scenarios. In particular, in some situations one may want to limit the sets $S$, $F$, and $P$ in the previous example to be of size $1$. However, it is clear that there are at least two factors at play which need further consideration:

\begin{figure}[t]
\begin{mdframed}[frametitle={%
        \tikz[baseline=(current bounding box.east),outer sep=0pt]
        \node[anchor=east,rectangle,fill=green!20]
        {\strut Collusion Golden Power Check};},linecolor=green!20,linewidth=2pt,frametitleaboveskip=\dimexpr-\ht\strutbox\relax]

\begin{tabular}{l p{9.25cm}}
\emph{Goal}: & The goal is checking whether an acquisition (of shares, stocks, etc.) causes any strategic Italian company to be possibly controlled by a set of foreign companies acting in collusion.
 \\[.4em]
\emph{Setting}: & Let $S$ be a set of strategic companies and $F$ be a set of foreign companies. Let $t$ be a transaction (e.g., an offer issued by a company $x$ to buy an amount $s$ of shares of a company $y$), with $x,y \in S \cup F$. \\[.4em]
\emph{Question}: \protect{   } & Decide whether $t$ causes $F$ to jointly control a company in $S$.\\[.4em]
\emph{Insight}: \protect{   } & Consider the possibility to 
exert Golden Powers to block $t$.
\\
\end{tabular}
\end{mdframed}

\vspace*{.5em}

\centering
  \includegraphics[width=0.75\textwidth]{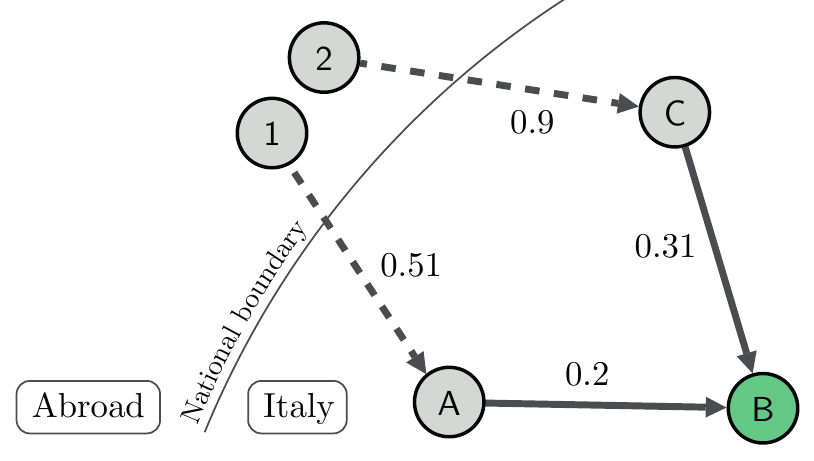}

\caption{Definition (top) and example (bottom) for Collusion Golden Power Check.}
\label{fig:collusion}
\end{figure}

\begin{enumerate}
    \item \emph{collusion} between companies, i.e., two or more companies act together even though there is no formal control established between them.
    \item \emph{missing} or \emph{unknown} information, i.e., ownership that is present in reality is not represented in the Knowledge Graph.
\end{enumerate}

\begin{figure}[t]
\begin{mdframed}[frametitle={%
        \tikz[baseline=(current bounding box.east),outer sep=0pt]
        \node[anchor=east,rectangle,fill=green!20]
        {\strut Cautious Golden Power Check};},linecolor=green!20,linewidth=2pt,frametitleaboveskip=\dimexpr-\ht\strutbox\relax]

\begin{tabular}{l p{9.25cm}}
\emph{Goal}: & The goal is checking whether an acquisition (of shares, stocks, etc.) causes any strategic Italian company to be possibly controlled by a foreign company, for which
shareholding information is incomplete. 
 \\[.4em]
\emph{Setting}: & Let $S$ be a set of strategic companies and $F$ be a set of foreign companies. Let $t$ be a transaction (e.g., an offer issued by a company $x$ to buy an amount $s$ of shares of a company $y$), with $x,y \in S \cup F$. \\[.4em]
\emph{Question}: \protect{   } & Decide whether $t$ causes any company in $f \in F$ to control a company in $S$, assuming that any missing ownership is in fact owned by $f$.\\[.4em]
\emph{Insight}: \protect{   } & Consider the possibility to 
exert Golden Powers to block $t$.

\end{tabular}
\end{mdframed}

\vspace*{.5em}

\centering
  \includegraphics[width=0.75\textwidth]{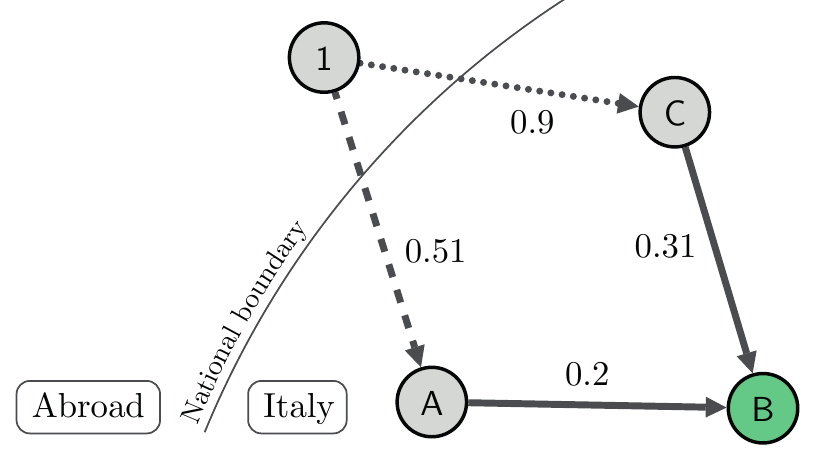}

\caption{Definition (top) and example (bottom) for Cautious Golden Power Check.}
\label{fig:cautious}
\end{figure}

\medskip
\noindent
\textbf{Collusion Golden Power Check}.
We first formulate the \emph{Golden Power Check} problem with the possibility of collusion, which we call \emph{Collusion Golden Power Check}.
It is clear that the scenario can be extended to \emph{Golden Power Limit} and \emph{Golden Power Protection} as well. However, it is also clear that it can be extended based on the degree of possibilities to be considered. For example, should it be considered that companies outside the set $F$ (both foreign and Italian companies) are potentially acting in collusion as well? This can easily be handled by extending the set $F$ to all companies considered to be potentially acting in collusion. At the same time, it is also clear that the act of choosing the set $F$ is in itself important. The definition and an example are given in Figure~\ref{fig:collusion}.

\medskip
\noindent
\emph{Example}. Consider the example given in the lower part of Figure~\ref{fig:collusion}. We observe a setting similar to the one in Figure~\ref{fig:goldenpowercheck}, discussing Golden Power Check. Transaction $t_1$ proposes for company $1$ to obtain $51\%$ of $A$, and transaction $t_2$ proposes for company $2$ to obtain $90\%$ of $C$. Both transactions fulfill the Golden Power Check, as no company under investigation by itself obtains control of strategic company $B$. On paper, given that $1$ and $2$ do not have any formal ownership relations between them, it can be assumed that this is not an issue. However, assuming that companies $1$ and $2$ may have some hidden form of control or other form of collusion between them, this may very well be a problem, and could be investigated for potential use of Golden Power. This concludes the example.

The definition of Figure~\ref{fig:collusion} can be effectively formulated as the following \vadalog reasoning task, which extends the definition we have seen before by adding (1) below:
\begin{align*}
  \textit{F}(x), \textit{F}(y) & \to \textit{Control}(x,y) \tag{1} \\
  \textit{T}(x,y,w) & \to \textit{Own}(x,y,w) \tag{2} \\
  \textit{Control}(x,y) & \to  \textit{Control}(x,x)    \tag{3} \\
  \textit{Control}(x,y), \textit{Own}(y, z, w), v = \textit{msum}(w, \langle y \rangle), v>0.5 &\rightarrow  \textit{Control}(x,z)   \tag{4} \\
  \textit{F}(x), \textit{S}(y), \textit{Control}(x,y) & \to \textit{CGPCheck}(x,y) \tag {5}
\end{align*}

\noindent
We have already discussed Rules 2 to 5 when considering \emph{Golden Power Check}. Rule 1 encodes the intuition that companies in $F$ may be in collusion, i.e., effectively exerting control over each other.

\medskip
\noindent
\textbf{Cautious Golden Power Check}.
The second problem considered above is missing or unknown information. In our examples, and in our real-world Knowledge Graph, if 80\% of the ownership of a company is represented, do we assume that the remaining 20\% are not relevant to our problem? This is what we considered in the problems so far, and is similar to what in many settings is called a ``closed world'' assumption. However, in a \emph{cautious} scenario, we may need to assume the most critical case, namely that the remaining 20\% are under the control of the companies under investigation. We thus introduce a \emph{cautious} variant of our problem in Figure~\ref{fig:cautious} and call it \emph{Cautious Golden Power Check}.

\medskip
\noindent
\emph{Example}. Consider the example given in the lower part of Figure~\ref{fig:cautious}.  Again, we observe a similar setting as the one in Figure~\ref{fig:goldenpowercheck} discussing Golden Power Check. However, the dotted line expressing that $1$ owns $90\%$ of $C$ expresses that this information is \emph{not} present in our Knowledge Graph, i.e., unknown. Similarly critical, the remaining $49\%$ of ownership of $B$ are unaccounted for, hence in the worst case, i.e., being \emph{cautious} we must assume they are under direct or indirect control of $1$. It is clear that many variations of the degree of \emph{caution} are possible here beyond the one defined in Figure~\ref{fig:cautious}. This example and definition should give a hint towards the factor of how unknown or missing information can be handled.  This concludes our example. 

\medskip
\noindent
Clearly, this problem can be combined with the two other base scenarios (\emph{Golden Power Limit} and \emph{Golden Power Protection}) as well as with the \emph{Collusion} variant. While we believe that the two variants considered here, namely \emph{Collusion} and \emph{Cautious} reasoning, are the most critical, it is clear that a number of variations are possible and deserve investigation.

\medskip
\noindent
\textbf{Summary}. In this section we explored three fundamental problems on (i) detecting possibly hidden takeovers through a \emph{Golden Power Check}, (ii) advising companies on the limits of transactions that can be made without having to invoke Golden Power through \emph{Golden Power Limit}, and (iii) giving options for proactively protecting companies through \emph{Golden Power Protection}. We also gave two advanced variants of these three fundamental KG-based reasoning tasks in (iv) considering collusion between companies and (v) considering missing and incomplete information.

Many other relevant KG-based applications and reasoning problems exist, and this section's two main messages are presenting a very relevant class of problems that can be solved based on the KG construction and analytics techniques presented in this work, and giving general advice on which kinds of advanced problems can be solved using an advanced Knowledge Graph.

\section{Related Work}
\label{sec:relwork}
There is a variety of literature that can be considered related to this work. 

\medskip
On the economic and statistical perspective, studies concerning the impact of the crisis on the different types
of economic activity are appearing. For example, a study of the regional differences in Austria w.r.t.\ economic vulnerability has been conducted~\cite{WIFOCoronaCrisisRegionalDiff}. 
They adopt the European \emph{National Classification of Economic Activities} (NACE code) and evaluate  impact on a five-level scale.
Unlike our work, they do not take into consideration the network dimension in the analysis.
As a further difference, we do not aim at analyzing the impact on economic terms, but provide
Knowledge Graphs, with a set of standard and advanced descriptive indicators, so as to complement and
favor economic analyses. We adopt the Italian  hierarchical classification of economic activities (called ATECO, last updated in 2007) 
issued by the Italian National Statistical Office~\cite{ISTATAteco}. 

\medskip
Studies on the topology of company networks are not new and the corporate economics community has been studying
it at various levels~\cite{Alme06,Barca01,Franks95,glattfelder2009backbone,Grano95,Porta99}, in order to
discuss dispersion of company control and define indicators. These works do not adopt a reasoning or algorithmic approach, but tend
to privilege a matrix-based formulation, which often exhibits non-trivial computational limitations. 
Moreover, they often rely on sampling strategies. Our reasoning-based approach
for the calculation of relevant indicators overcomes computational problems thanks to the adoption of a tractable 
(i.e., polynomial time) language for Knowledge Representation and Reasoning (KRR) and at the same time
leverages the Italian company network.

\medskip
In this work we introduce the novel notion of $\epsilon$-conglomerate (Definition~\ref{def:conglomerate}), which
turns out to be useful to analyze the structure of the company graph.
It is based on a binary relation of vicinity (Definition~\ref{def:same-conglomerate-c-close}),
defining how close two nodes of the KG are w.r.t. integrated ownership.
Vicinity generalizes and captures the concept of \emph{close links},
adopted in the context of \emph{collateral eligibility} as defined by the European Central Bank regulations~\cite{ECB14}.

\medskip
The Artificial Intelligence and Computer Science communities have shown special theoretical
interest towards problems related to company graphs. For example, Ceri et al.~\cite{datalog1}
first define the notion of company control in logic programming terms and Romei et al.~\cite{Rome15}
propose an algorithmic approach to compute company ownership; however, their approach is not adaptable
to the COVID-19 case, as it does not rely on a declarative high-level specification, but offers static
implementations, addressing one specific problem.

\medskip
We have recently adopted company ownership and company control problems as reference
use cases of novel KRR formalisms for reasoning in KGs~\cite{BGPS17} and explicitly developed applications of
the Vadalog System to address industrially relevant problems related to company KGs~\cite{ABIS20}.
To the best of our knowledge, the only ongoing COVID-19 research work partially leveraging reasoning on KGs
pertains to literature organization, with a KG built by Oelen~\cite{KG4CovidResearch}, trying to gather
all the existing research contributions, their attributes and relationships.

\section{Future Work, Challenges and Perspectives}
\label{sec:conclusions}
With this paper we aimed at contributing a visionary opinion about
the application of KG technologies to conduct impact analysis of
the COVID-19 pandemic, with special reference to evaluating how
it affects the structure of economic networks in the Italian company graph.
The context has been introduced with a thorough explanation of our
main working models, including integrated ownership and company control,
which are helpful to reason on conglomerates and protection of strategic companies,
respectively. In particular, our novel formalization of integrated ownership
allows to unambiguously define a company conglomerate and therefore study
its properties; similarly, leveraging company control, we formulated
a set of problems encoding the scenarios that arise in the protection
of strategic companies from takeovers.

\medskip
Logic-based knowledge graphs are an ideal means to formulate problems
in this realm and modern KRR languages such as \vadalog allow for
direct execution on automated-reasoning systems. Nevertheless,
building a real-world KG and activating reasoning tasks upon them
is by no means trivial and requires a principled combination of
AI technologies, data integration activity and ad-hoc algorithm development,
which we carefully conducted, but whose details are outside the scope of this paper.

\medskip
This contribution should be considered as a work in progress,
in the sense that as the COVID-19 emergency is swiftly moving
with frequent decrees and local regulations being released,
we expect to produce, discuss and make available upon request
more KGs, applying and refining the discussed techniques.
Moreover, multiple aspects of great interest are still open and will be
matter of work in the next future. In detail, in this work we have
considered the KG of non-listed Italian companies, but we plan to
extend the KG at our disposal by including listed companies.
Geographical regulations have been considered up to the
regional level. Yet, we expect a growth of city-level
amendments or ad-hoc waivers for specific companies, which are
not captured by our analysis. Unfortunately, the respective data
is not available and probably not even yet digitalized. We expect
the situation to rapidly change and we will integrate such information
in the provided KGs.

\medskip
Conglomerates are a powerful tool to analyze the structure of company groups.
In this work, we decomposed the Italian network into groups
identified by the presence of share majority ownership chains. It is our plan to generate
conglomerate breakdowns for the KGs at hand considering at least
two more thresholds, so as to encompass a looser notion of conglomerate
including companies with mild financial relationships as well as
tighter ones, where multiple companies
actually operate as a single center.

\medskip
Conglomerates can be used for more sophisticated analyses
whereas so far we have presented only basic descriptive analytics,
anyway impossible to compute without the conglomerate breakdown.
Specific indicators may consider the dispersion resp. concentration
of control in a specific company group; this would provide
an additional feature to assess the group robustness. Furthermore,
conglomerates are interesting per se. Each has a peculiar topology
that we want to study with both quantitative methods based on
well-known network analysis indicators and reasoning-based techniques,
to assess the presence of recurring proprietary structures, e.g.,
capturing them via KG embeddings, a rising scientific field in the AR community.

\medskip
Italian Golden Power tools exemplify contexts where KGs are
extremely helpful for the evaluation of evolving scenarios in
graph-based domains. We plan to develop 
specific KG-based services to support decisions
to protect strategic companies in crisis scenarios. 

\medskip
We will report about all these aspects in upcoming versions of this work.

\bigskip
\bigskip
\noindent
{\small
\textbf{Acknowledgements}. E.\ Sallinger acknowledges the support of the WWTF (Vienna Science and Technology Fund) grant VRG18-013, the EPSRC grant  EP/M025268/1, and the EU Horizon 2020 grant 809965.
}

\bibliographystyle{myplain}
\bibliography{b}	

\end{document}